\documentclass[10pt,twocolumn,letterpaper]{article}

\usepackage{iccv}              

\definecolor{iccvblue}{rgb}{0.21,0.49,0.74}
\usepackage[pagebackref,breaklinks,colorlinks,allcolors=iccvblue]{hyperref}

\usepackage{tabularx}
\usepackage{multirow}

\def\paperID{4290} 
\def\confName{ICCV}
\def\confYear{2025}

\title{SCHNet: SAM Marries CLIP for Human Parsing}

\author{Kunliang Liu$^{1,2\textsuperscript{\dag}}$, Jianming Wang$^{2}$, Rize Jin$^{2}$, Wonjun Hwang$^{3*}$, and Tae-Sun Chung$^{1*}$\\
$^{1}$Ajou University, Korea, $^{2}$Tiangong University, China, $^{3}$Korea University, Korea\\
{\tt\small \{tjpulkl, tschung\}@ajou.ac.kr, \{wangjianming, jinrize\}@tiangong.edu.cn, wjhwang@korea.ac.kr }
}

\begin{document}
\maketitle

\begingroup
\renewcommand\thefootnote{}
\footnotetext{$^*$Corresponding authors and $\textsuperscript{\dag}$this paper was jointly completed by Tiangong University and Ajou University, which are acknowledged as co-first contributing institutions.}
\endgroup

\begin{abstract}
 Vision Foundation Models (VFM) such as the Segment Anything Model (SAM) and Contrastive Language-Image Pre-training Model (CLIP) have shown promising performance for segmentation and detection tasks. However, although SAM excels in fine-grained segmentation, it faces major challenges when applying it to semantic-aware segmentation. While CLIP exhibits a strong semantic understanding capability via aligning the global features of language and vision, it has deficiencies in fine-grained segmentation tasks. Human parsing requires to segment human bodies into constituent parts and involves both accurate fine-grained segmentation and high semantic understanding of each part. Based on traits of SAM and CLIP, we formulate high efficient modules to effectively integrate features of them to benefit human parsing. We propose a Semantic-Refinement Module to integrate semantic features of CLIP with SAM features to benefit parsing. Moreover, we formulate a high efficient Fine-tuning Module to adjust the pretrained SAM for human parsing that needs high semantic information and simultaneously demands spatial details, which significantly reduces the training time compared with full-time training and achieves notable performance. Extensive experiments demonstrate the effectiveness of our method on LIP, PPP, and CIHP databases. 
\end{abstract}

\begin{figure}[t]
\centering
\includegraphics[width=1.0 \linewidth]{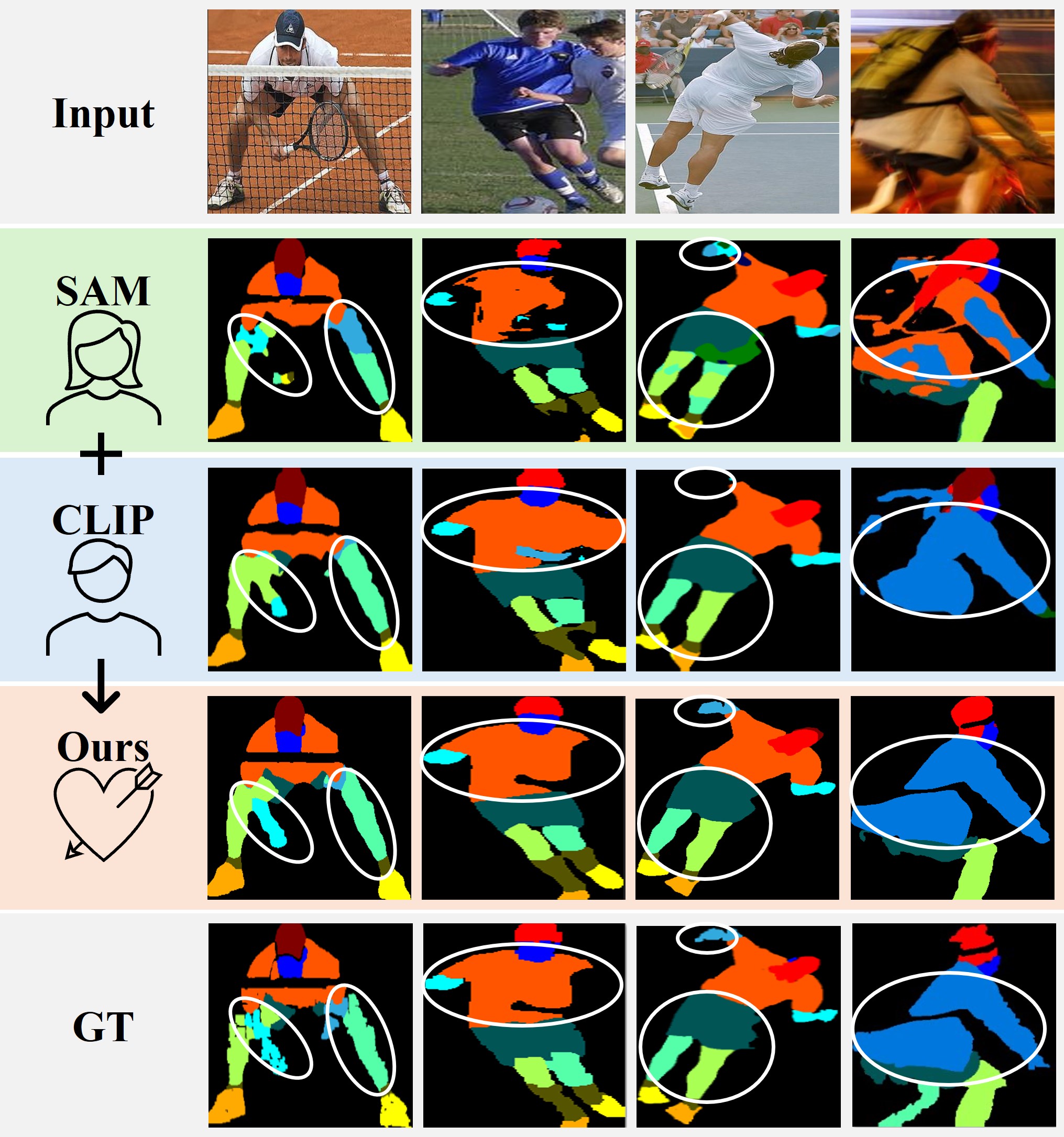}
\vspace{-0.6cm}
\caption{\textbf{Visual comparison among SAM, CLIP, and our SCHNet} on human parsing. The regions highlighting the differences are marked with white circles. SAM make fine-grained segmentation without missing regions, but its outputs tend to be noisy, while CLIP provides coarse predictions, sometimes failing to parse entire important parts. Our method integrates the strengths of both methods, ensuring stable and reliable performance.
}
\vspace{-0.4cm}
\label{fig:Comp_results}
\end{figure} 

\section{Introduction}
\label{sec:intro}
Human parsing aims to decompose a human in an image into constituent parts, including different human body parts and clothing items (e.g., arms, legs, coat, dress, etc.). Moreover, the distinction between the left and right of human body parts is also necessary, such as the left and right arms. Human parsing belongs to the subfield of scene parsing, but it is more complicated than scene parsing due to the deformable poses of persons, the intricate textures and styles of clothing, and the variation in the size of various parts of the human body. Therefore, networks that perform accurate human parsing must effectively integrate not only the fine details but also the high-level semantic features of the input images. Accurate human parsing is critical for human understanding and various human-centric applications such as virtual try-on and human-robot interactions.

After the emergence of fully convolutional network (FCN)~\cite{Jonathan2015FCN-13}, many methods~\cite{zhao2017pspnet,Liang2015,LiangChen2018,hu2018senet,rao2022dynamicvit,li2022deep,dai17dcn,replknet,LiuHwang2022,CSENet2024,HssN2024} have been developed to improve human parsing performance. Vision transformer-based networks~\cite{Alexey2021,Nicolas2020,wang2021not,rao2021dynamicvit,Xia_2022_CVPR,liu2021Swin,zhang2022topformer,yu2021metaformer,yuan2022volo,ci2023unihcp,tang2023humanbench}, and hybrid networks~\cite{rao2022dynamicvit,pan2021integration,yan2021contnet,SSUCSFR2024,yang2024humanparsing} that integrate convolution with the transformer are also proposed. All of these networks significantly enhanced the ability to encode fine details and high-level semantic features and consequently harvested the impressive performance. Some other methods that relied on the hierarchical structure of the human body~\cite{HssN2024}, the distribution rule of each human part~\cite{LiuHwang2022} and the keypoint information of the person ~\cite{Zhang2020CorrPM-8} also achieved improved results. Distinct human-centric tasks may benefit each other. In view of this, there are also some attempts to train a shared neural network for different tasks jointly~\cite{tang2023humanbench,ci2023unihcp}, such as training human parsing in conjunction with human keypoint detection, pedestrian attribute recognition, or person re-identification. Although previous methods have achieved great success on human parsing, they are all trained from scratch, which not only require excessive training time but also demand much more annotations for training. 
Moreover, multi-modal models that integrate information from multiple modalities have made remarkable progress and emerged as one of the most active areas of study~\cite{CLIPSAM,clip2021,koleilat2024medclip,clip_as_rnn,MedCLIP}. Similarly, human parsing can benefit from leveraging multiple modalities, e.g, textual descriptions and images, to reduce training time, mitigate overfitting, and improve overall performance.

In this paper, we integrate SAM~\cite{sam2023} with CLIP~\cite{clip2021} for human parsing. In this way, we can combine multi-modal information, utilize the pretrained weights of the Vision Foundation Models (VFMs) and consequently accelerate the training process and notably improve the performance. CLIP shows a strong semantic understanding ability of different modalities via aligning the global features of language and vision. However, CLIP-based methods struggle with precise image parsing, making them unsuitable for direct fine-grained segmentation~\cite{CLIPSAM}. As we can see from the third row in~\cref{fig:Comp_results} that CLIP produces insufficient segmentation. While SAM pre-trained on a large corpus of images excels in fine-grained segmentation and generalizes across various vision domains, there still exist major challenges in applying SAM to semantic-aware segmentation tasks. As the second row of~\cref{fig:Comp_results} views that SAM always generates over-segmenting results. To effectively employ the advantages of SAM and CLIP for human parsing, we formulate two modules, the Semantic-Refinement Module (SRM) and Fine-tuning Module (FTM). Unlike other methods~\cite{CLIPSAM,koleilat2024medclip,clip_as_rnn} that employ CLIP output as prompt constraints of SAM for precise semantic parsing or as rough results for post-processing, we formulate SRM to combine the output features of CLIP in each stage with the corresponding features in each stage of SAM. In this way, we fully fuse multiple levels of semantic information of CLIP with multiple stage details of SAM to substantially improve the parsing performance. Inspired by Rein~\cite{Wei_2024_CVPR} and Adapter~\cite{chen2023sam}, we design the FTM module to add trainable tokens to each layer feature maps and partly fine-tuning the pre-trained weights of each layer of SAM to enhance the representation ability of output features.

The major contributions of our work are as follows.
\begin{itemize} 
    \item We introduce a Semantic-Refinement Module where we first improve the semantic representation ability of CLIP features and then integrate them with each corresponding stage feature of SAM to significantly improve the semantic-aware parsing ability of SAM.
    \item We propose a Fine-tuning Module to append learnable tokens to each layer feature maps and partly fine-tune the pre-trained weights of SAM. The formulated module improves the parsing performance, accelerates the convergence of the network, and shortens the training time.
    \item We demonstrate the significant performance improvement and much shorter training time gained by the proposed method over well-known human parsing methods through extensive experiments on Look into Person (LIP), Pascal-person-Part (PPP), and Crowd Instance-level Human Parsing (CIHP) benchmark. 
\end{itemize}

\section{Related Works}
\textbf{Human Parsing:} 
Human parsing is a fine-grained semantic segmentation task in which all pixels of the human image are labeled. As a subfield of parsing tasks, human parsing is more challenging. To obtain a reasonable parsing result, the networks that perform human parsing should not only focus on fine details to correctly categorize each pixel in the image, especially the boundaries among different parts, but also concentrate on strong semantic information to distinguish each human part, particularly those parts that have similar textures and appearance but should be categorized into left and right, such as arms, legs, and shoes. Many deep learning-based methods~\cite{Ruan2019CE2P-21,DTML2024,CSENet2024,LiuHwang2022} have gained significant improvements in human parsing through capturing multi-stage details. Many other methods have achieved impressive results by learning additional tasks~\cite{tang2023humanbench,ci2023unihcp}, combining with additional annotations (such as keypoints)~\cite{Zhang2020CorrPM-8} or constructing hierarchical structure of the human body~\cite{XiaoPCNet2020,HssN2024,Wang2020HTypePRR-20} to achieve contextual or high-semantic information. All these methods are not multi-modal networks that can effectively leverage easily accessible description information to benefit parsing. Moreover, these models are trained from scratch, which requires a substantially long training time and much larger datasets. Recently, there has been an increasing focus and attention on foundation models. Various foundation models have achieved remarkable performance on downstream tasks~\cite{CLIPSAM,koleilat2024medclip,clip_as_rnn,yang2023foundation,chen2023sam,Wei_2024_CVPR,sam_hq,MedCLIP}. Based on foundation models, multi-modal data can be easily integrate with the training data to accelerate the training process and achieve more robust, and higher performance. However, there is yet no method that utilizes the foundation model for human parsing. 

\begin{figure*}[t]
\centering
\includegraphics[width=1.0\linewidth]{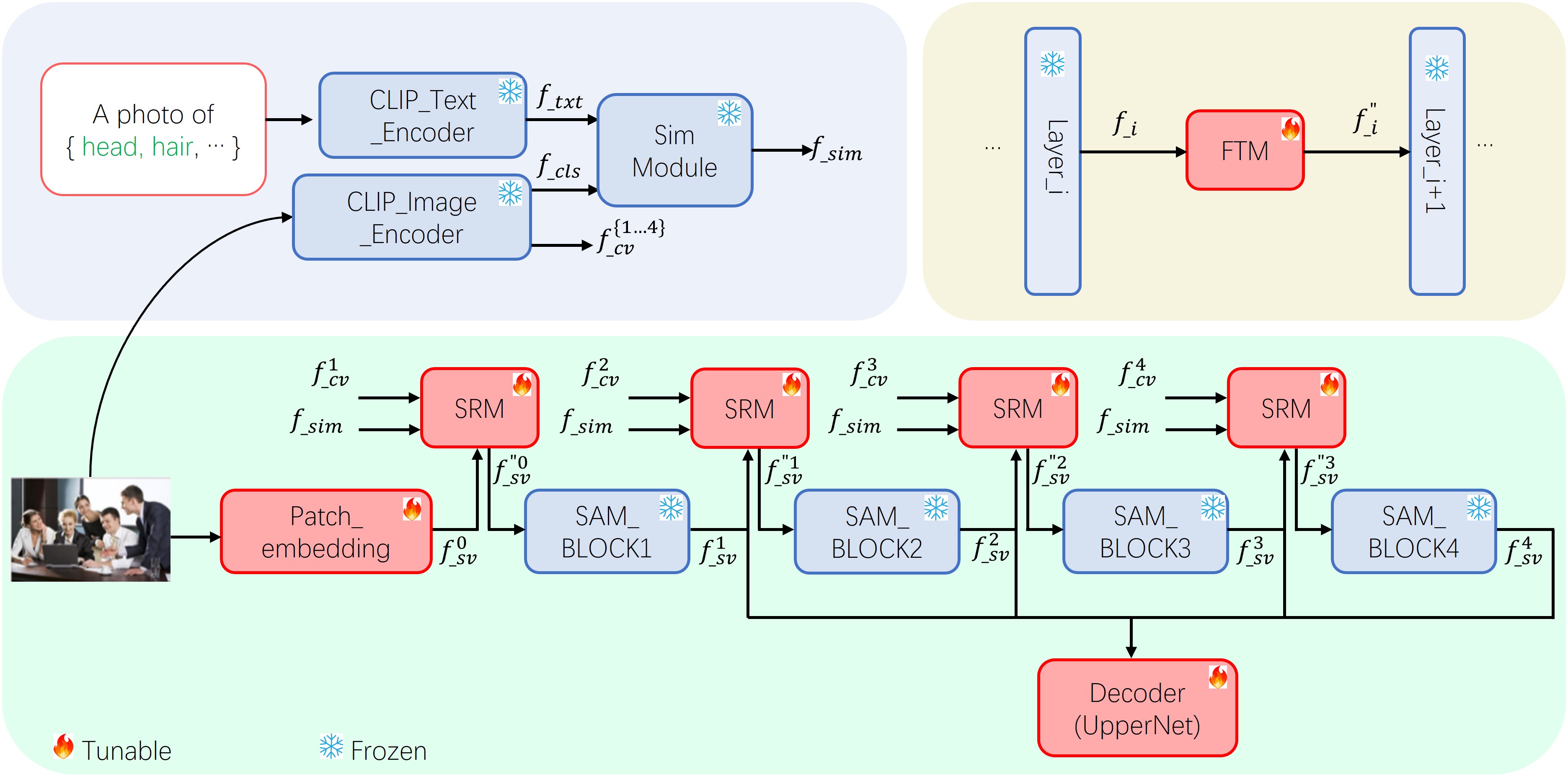}
\vspace{-0.2cm}
\caption{\textbf{Architecture of SCHNet}. $f_{\_txt}$: the text feature output by pre-trained text encoder of CLIP. $f_{\_cls}$: the class embedding feature output by pre-trained image encoder of CLIP. $f_{\_cv}^{\{1...4\}}$: the feature maps output by pre-trained image encoder of CLIP. We leverage feature maps of all blocks (from $1$ to $4$) of image encoder of CLIP. $f_{\_sim}$: the Similarity feature that is calculated from text feature and class embedding feature using $SimModule$. $Layer\_i$ and $Layer\_i+1$: means the $ith$ and $(i+1)th$ layers of SAM network.  $f_{\_i}$ means the output feature maps of $ith$ layer of SAM. $f_{\_i}^{"}$: means the fine-tuned feature maps utilizing $FTM$. $f_{\_sv}^{0...4}$: means the output feature maps after patch embedding and $4$ blocks of SAM. $f_{\_sv}^{"0...4}$: means the semantic strengthened feature maps after SRM. We combine each stage semantic information of CLIP with each stage feature maps of SAM to improve the semantic-aware segmentation performance of pre-trained SAM fine-tuned by FTM module.
}
\label{fig:Network}
\vspace{-0.4cm}
\end{figure*}

\noindent\textbf{CLIP and SAM Models:} 
Obviously, CLIP~\cite{clip2021} and SAM~\cite{sam2023} are two representative models with robust and impressive performance in classification and segmentation tasks. CLIP aims to understand image content and match it with the corresponding natural language description by training on a dataset of about $400$ million image-text pairs collected from the Internet. By aligning multi-modal features, CLIP possesses robust semantic understanding abilities for both language and vision. However, CLIP has deficiencies in the fine-grained segmentation task. SAM establishes a general foundation model for image segmentation through training on $11$ million high-resolution images and $1.1$ billion high-quality segmentation masks. While SAM excels in fine-grained segmentation, it has limited semantic-aware ability and tends to generate numerous redundant masks and accordingly requires complex post-processing. Therefore, many methods~\cite{salip2024,CLIPSAM,koleilat2024medclip,clip_as_rnn, yang2023foundation} proposed to integrate CLIP and SAM together to harness the advantages of the two foundation models. However, the majority of these methods~\cite{CLIPSAM,koleilat2024medclip,clip_as_rnn,yang2023foundation} utilize CLIP to provide SAM with extra prompts (such as rough results in the form of points, boxes, etc.) based on the final output feature of CLIP to enhance its ability of accurate segmentation. Extra prompts demand post-process, which reduces the model efficiency and generality. Moreover, merely relying on the output of the final stage of CLIP can not fully exploit each stage semantic information of CLIP. Our SRM module employs text feature, class embedding, and each stage feature of CLIP to inject multiple level semantic information to each SAM stage, which considerably benefits human parsing. Post-process is no longer required. The proposed FTM module adds learnable tokens to the feature maps of each layer of SAM and constructs a shared structure that has the capability of fine-tuning the output feature of SAM to shorten training time and enhance performance.

\section{Methodology}
Sine human parsing is more challenging than general purpose semantic segmentation, it highly demands both fine-grained details and high level of semantic understanding of each human part. In this session, we propose an effective method to exploit the advantages of two VFMs (e.g., SAM and CLIP) for efficient training and remarkable performance of human parsing. Emphatically, the main motivation of this paper is how to transit from the general purpose VFMs to the human parsing successfully. For this purpose, we formulate SRM and FTM. 
The SRM module employs an effective structure to instill strong semantic information that is derived from pre-trained CLIP to enhance the semantic understanding ability of SAM. 
The FTM module adds learnable information to SAM feature maps to transfer the feature maps that focus on domain generalized information to the specific human parsing domain. 
Moreover, in the proposed module, we design a highly efficient squeezing and expanding mechanism to fine-tune the pre-trained output feature maps of SAM. 
Through utilizing the aforementioned modules, the feature maps that input the decoder contain not only adequate fine-details but multi-level of semantic understanding information, which remarkably improve the human parsing performance under much shorter training time compared with scratch training. 
\cref{fig:Network} illustrates the overall network structure.

\begin{figure}[t]
\centering
\includegraphics[width=1.0 \linewidth]{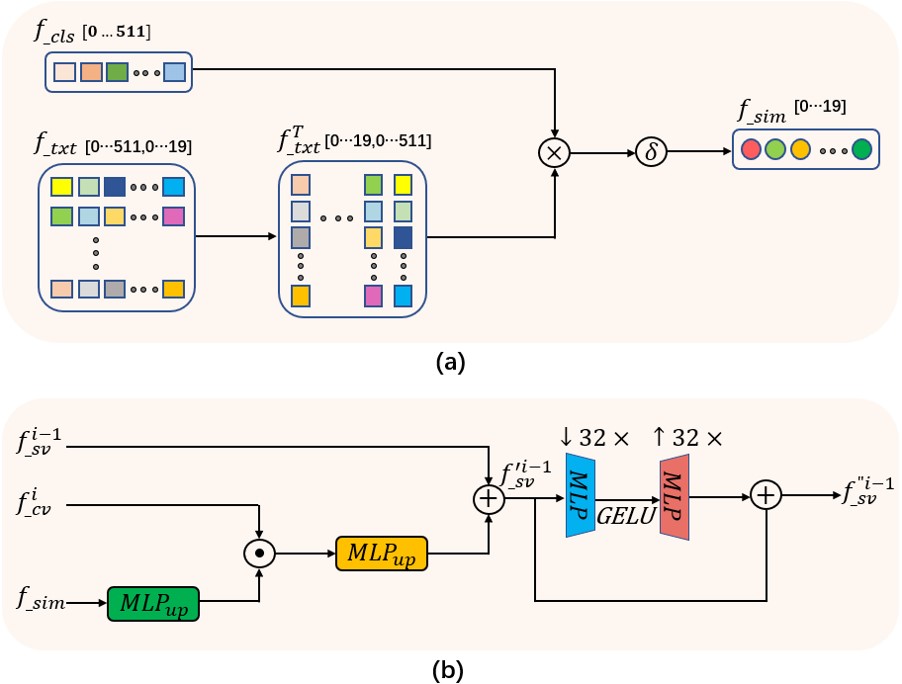}
\vspace{-0.6cm}
\caption{\textbf{Overview of SimModule and SRM}. (a) SimModule structure, (b) SRM module structure. $f_{\_cls}^{\{0...511\}}$: means class embedding feature output by CLIP image encoder, $\{0...511\}$ is the dimension range of class embedding feature. $f_{\_txt}^{\{0...511,0...19\}}$: is the text feature output by text encoder of CLIP, $\{0...511,0...19\}$ is the dimension range of the text feature. We employ the LIP dataset as an example. In LIP dataset, there exits $20$ categories of human parts. $\bigotimes$: matrix multiplication, $\bigoplus$:element-wise addition,  $\bigodot$: element-wise multiplication, \textcircled{$\delta$}: Softmax activation. $f_{\_sv}^{i-1}$: means feature maps from $(i-1)th$ stage of SAM.$f_{\_cv}^{i}$: means feature maps from $ith$ stage of CLIP. $f_{\_sim}$: means similarity between text and class embedding feature. $\uparrow$ and $\downarrow$: mean increase and decrease the channel dimension to what times of input dimension.
}
\vspace{-0.4cm}
\label{fig:SRM}
\end{figure} 

\begin{figure}[t]
\centering
\includegraphics[width=1.0 \linewidth]{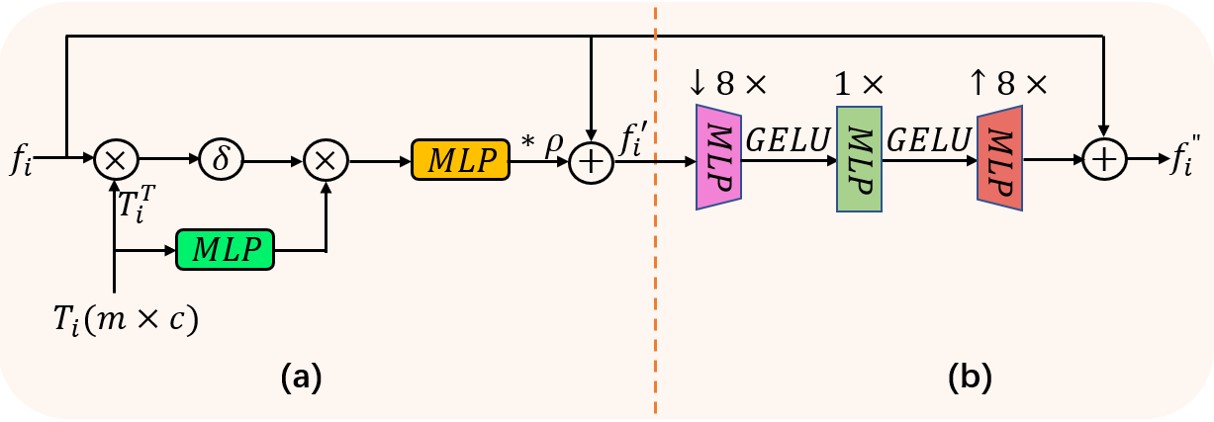}
\vspace{-0.6cm}
\caption{\textbf{Overview of FTM}. (a): module that is used to add learnable token information. (b): module that is leveraged to fine-tune the feature maps of SAM. $f_{i}$,$f_{i}^{'}$,$f_{i}^{"}$: mean feature maps from $ith$ layer of SAM, feature maps added with learnable tokens information, and fine-tuned feature maps that input to next layer of SAM, respectively. $T_{i}^{T}$:means the transposed $ith$ learnable tokens, $m\times c$ means the token dimension. $\bigotimes$,$\bigoplus$, \textcircled{$\delta$}: mean matrix multiplication, element-wise addition, and Softmax activation, respectively. $\rho$ is a learnable parameter. $*$ means multiplication by a coefficient.
}
\vspace{-0.4cm}
\label{fig:FTM}
\end{figure} 

\subsection{ Semantic-Refinement Module (SRM) }
We employ the image encoder $Enc_{\_cv}$ of the pre-trained CLIP model~\cite{clip2021} to extract class-embedding and vision features, $f_{\_cls},f_{\_cv}^{1\sim 4}=Enc_{\_cv}(I)$, from the input image $I$. Here, $1\sim 4$ means that we obtain multiple features from the blocks $3$, $5$, $7$, and $11$. We first generate a text prompt T from labels of the human parsing dataset (such as LIP). The text encoder $Enc_{\_ct}$ of CLIP is utilized to extract textual features, $f_{\_txt}=Enc_{\_ct}(T)$, of input text prompts $T$. As part $(a)$ of ~\cref{fig:SRM} views, we apply matrix multiplication to the class embedding and the transposed textual feature to calculate the similarity between them. After regularizing the similarity using $Softmax$, we achieve the possibility of each category exist. The equation is denoted as: 

\begin{equation}
    f_{\_sim} = Softmax( f_{\_cls} \otimes f_{\_txt}^{T} ),
    \label{eq:Sim}
\end{equation}
where $f_{\_sim}$ is the similarity vector between the textual feature and class embedding feature. $Softmax$ means $Softmax$ activation. $f_{\_cls}$ is class embedding. $\otimes$ means matrix multiplication. $f_{\_txt}^{T}$ means the transposition of the matrix $f_{\_txt}$ and $f_{\_txt}$ is the textual feature from text encoder of CLIP. 
The dimension of $f_{\_sim}$ equals to the number of class number in the dataset, for instance, the number equals $20$ in LIP dataset. As the left part of (b) in ~\cref{fig:SRM} showcases, we increase the dimension of $f_{\_sim}$ to match the dimension of vision feature of CLIP by utilizing one Multilayer Perceptron ($MLP$). Then, we perform element-wise multiplication of the similarity feature with the feature map generated by the CLIP image encoder. By this means, we enhance the class recognition ability of image feature maps from all stages of the CLIP image encoder. Through another $MLP$, we further increase the dimension to have the same dimension number as feature maps that come from SAM image encoder. Finally, we combine the feature maps from CLIP with the feature maps from SAM via element-wise addition operation to inject the semantic-aware information to SAM's features. The process can be denoted by the following equation;
\begin{equation}
    f_{\_sv}^{'i-1} = MLP(f_{\_cv}^{i} \odot MLP(f_{\_sim})) \oplus f_{\_sv}^{i-1},    
    \label{eq:srm}
\end{equation}
where $f_{\_sv}^{'i-1}$ means the feature map that we achieved via injecting the multi-level semantic information, $i\in [1...5]$. $f_{\_cv}^{i}$ is the feature map from the $ith$ stage of CLIP image encoder. $MLP$ means Multilayer Perceptron used to increase the channel number of the feature, $\odot$ means element-wise multiplication, and $\oplus$ means element-wise addition, respectively. $f_{\_sv}^{i-1}$ derives from the $(i-1)th$ stage of SAM image encoder, as shown in ~\cref{fig:Network}. In our method, all $MLP$s are composed of the learnable weights $W$ and biases $b$. We denote it as the following formula:
\begin{equation}
 f_{out} = W \times f_{in} + b,
    \label{eq:mlp}
\end{equation}
where $f_{out}$ means the output feature map from $MLP$ and $f_{in}$ means the feature map that input to $MLP$.
As the right part of $(b)$ in ~\cref{fig:SRM} shows, to further reduce the redundant feature and focus on important details and semantic information, we exploit a similar mechanism to squeezing and expanding process. The equation is as following:
\begin{equation}
    f_{\_sv}^{"i-1} = f_{\_sv}^{'i-1} \oplus MLP(GELU(MLP( f_{\_sv}^{'i-1})),
    \label{eq:srm2}
\end{equation}
where $f_{\_sv}^{"i-1}$ is the achieved feature that we combine multi-level semantic information from CLIP with the feature from previous stage of SAM, then fine-tuned by two $MLP$s and further input to the next stage of SAM. $f_{\_sv}^{'i-1}$ is the result feature that we inject multi-level semantic information of CLIP to the feature from $(i-1)th$ stage. We use two $MLP$ to decrease and increase the dimension of the input feature map. We employ one $GELU$ activation between the two $MLP$s. Through leveraging SRM module, we can inject multi-level semantic information that from multiple stages of CLIP into the corresponding feature maps that from multiple stages of SAM. Furthermore, we employ class embedding and textual information to significantly enhance the semantic-aware ability of CLIP feature maps in SRM module and accordingly improve the semantic-aware capacity of SAM feature maps after semantic information injection. To maintain the efficiency, the SRM module is utilized in the shared manner within the network.  

\subsection{Fine-Tuning Module (FTM)}
Drawing inspiration from~\cite{Wei_2024_CVPR}, we employ the similar structure as Rein to add learnable tokens to each layer's feature map of SAM, as the top-right part of ~\cref{fig:Network} and the $(a)$ part of the FTM module in ~\cref{fig:FTM} illustrate. The learnable tokens are added to existing feature maps using the mechanism similar to attention that was adopted within neural networks. The learnable tokens can attach the learned specific domain knowledge, such as here the knowledge available to human parsing, to the general domain knowledge learned from pre-trained SAM. This mechanism can be denoted as:
\begin{equation}
    f_{i}^{'} = \rho * MLP(Softmax(f_{i} \otimes T_{i}^{T})\otimes MLP(T_{i}))\oplus f_{i},
    \label{eq:FTM1}
\end{equation}
where $\rho$ is a learnable parameter, $T_{i} \in T$, and  $T \in \mathbb{R} ^{n\times m\times c}$, $n$ means the layer number in SAM network, $m$ means token number, and $c$ means dimension of each token, $T_{i}^{T}$ is the transposed matrix of $ith$ layer's learnable tokens. As the (b) part of ~\cref{fig:FTM} showcases, to further fine-tune the feature after adding learnable tokens and continue to re-target the pre-trained general domain knowledge to the specific human parsing domain, we construct a multi-layer fine-tuning network module that composes of three $MLP$s, where the first and third $MLP$s are shared among all layers. The sharing mechanism can reduce the parameters to learn and improve the learning efficiency of the model. Our method leverages the first $MLP$ to decrease the channel dimension of the features. Through decreasing channel dimensions, we aim to reduce the redundant features and remove the interference features after attaching learnable tokens. The second $MLP$ is to fine-tune the features that are produced from SAM and the prior part of this module to adapt to the human parsing domain, and the third $MLP$ is to increase the dimension to align with the input feature. Two $GELU$ functions are appended after the first and second $MLP$, respectively. The formula is as follows:
\begin{equation} 
\resizebox{0.90\hsize}{!}{$
f_{i}^{"}=MLP(GELU(MLP(GELU(MLP(f_{i}^{'})))))\oplus f_{i} $},
    \label{eq:FTM2}
\end{equation}
where $f_{i}$ is the feature map from previous layer, $f_{i}^{'}$ means the feature maps with learnable tokens added according to a certain proportion, $f_{i}^{"}$ denotes fine-tuned feature maps using three projection layers. $GELU$ is the $GELU$ activation function. Because the learnable parameter $\rho$ begins from a very small value, drastic fluctuations of the additional information can be avoided, which benefits convergence of the pre-trained networks. Furthermore, the gradually added semantic information by SRM module substantially aids in additional information learning and pre-trained feature fin-tuning and eventually improves the human parsing performance.  

\section{Experiments}
We conduct comprehensive evaluations of our proposed method on three well-known benchmark datasets: LIP~\cite{Gong2017LIP}, PPP~\cite{PPP2017}, and CIHP~\cite{Gong2018PGN}.  We first introduce the datasets we used in our method. Then, we detail the implementation of the network, the setting of the training, and the inference. After that, we compared the state-of-the-art approaches with ours in terms of the Mean Intersection over Union (mIoU), Accuracy, Precision metrics, etc. 

\subsection{Datasets}
\noindent\textbf{Look In Person (\textbf{LIP}):}
The LIP dataset~\cite{Gong2017LIP} was utilized in LIP challenge 2016 for human parsing tasks. Totally, it consists of 50,462 single-person images with various resolutions gathered from real-world scenario, where 30,462 images are leveraged for training, 10,000 for validation, and 10,000 for testing. These images are captured from a wide range of viewpoints, occlusions, and complex backgrounds. All images are finely labeled at the pixel level with 19 semantic human part categories (including 6 body parts and 13 items of clothing) and one background class.

\noindent\textbf{PASCAL-Person-Part (\textbf{PPP}):} 
PPP dataset~\cite{PPP2017} is one of the most representative and widely utilized datasets in the human parsing task. It is the subset of the Pascal VOC dataset designed for human part analysis tasks. The images in the dataset offer a wide range of real-world scenario to support diverse research needs. There are multiple people appearances in an unconstrained environment. Each image comes with detailed annotations of human body parts. The dataset totally has $7$ classes, such as the head, torso, upper-arm, lower-arm, upper-leg, low-leg and background. There are 3,533 images in the dataset, where 1,716 for training and 1,817 for testing.

\noindent\textbf{Crowd Instance-level Human Parsing (\textbf{CIHP}):}
The CIHP dataset~\cite{Gong2018PGN} is a large-scale multi-person dataset that provides 38,280 images with pixel-wise annotation of 20 semantic parts, including the background. The images in CIHP are collected from a real-world scenario. Each image includes about three people, and the persons in the image appear with challenging poses and viewpoints, heavy occlusions, and show in a wide range of resolutions~\cite{Ke2019CIHP}. The dataset is elaborately annotated to benefit the semantic understanding of multiple people in the real situation and to enable a detailed analysis of semantic information in complicate multi-person scenes. The images in the dataset are divided into three sets, 28,280 images for training, 5,000 images for validation, and 5,000 images for testing. 

\begin{table*}[t]
\centering
\setlength\tabcolsep{2.6pt}
\resizebox{1.0\linewidth}{!}{%
\scriptsize
\begin{tabularx}{\linewidth}{p{1.4cm}<{\centering} c c c c c c c c c c c c c c c c c c c c c}
\toprule
Method 
&hat     &hair   &glove   &glass   &u-cloth &dress 
&coat    &sock   &pants   &j-suits &scarf   &skirt 
&face    &l-arm  &r-arm   &l-leg   &r-leg   &l-shoe 
&r-shoe  &bkg    &Avg    \\  
\midrule
SS-NAN~\cite{Zhao2017Self-SupervisedAgg}                       &63.86&70.12&30.63&23.92&70.27&33.51&56.75&40.18&72.19&27.68&16.98&26.41&75.33&55.24&58.93&44.01&41.87&29.15&32.64&88.67&47.92 \\
JPPNet~\cite{Liang2018JPPNet}                        &63.55&70.20&36.16&23.48&68.15&31.42&55.65&44.56&72.19&28.39&18.76&25.14&73.36&61.97&63.88&58.21&57.99&44.02&44.09&86.26&51.37 \\
CE2P~\cite{Ruan2019CE2P-21}                         &65.29&72.54&39.09&32.73&69.46&32.52&56.28&49.67&74.11&27.23&14.19&22.51&75.50&65.14&66.59&60.10&58.59&46.63&46.12&87.67&53.10 \\
SNT~\cite{Ji2020LSNT-22}                          &66.90&72.20&42.70&32.30&70.10&33.80&57.50&48.90&75.20&32.50&19.40&27.40&74.90&65.80&68.10&60.03&59.80&47.60&48.10&88.20&54.70\\
CorrPM~\cite{Zhang2020CorrPM-8}                       &66.20&71.56&41.06&31.09&70.20&37.74&57.95&48.40&75.19&32.37&23.79&29.23&74.36&66.53&68.61&62.80&62.81&49.03&49.82&87.77&55.33\\
SCHP~\cite{li2020self}                       &69.96&73.55&50.46&40.72&69.93&39.02&57.45&54.27&76.01&32.88&26.29&31.68&76.19&68.65&70.92&67.28&66.56&55.76&56.50&88.36&58.62\\
DTML~\cite{DTML2024}&68.07&73.86&43.62&34.27&\textbf{75.23}&\textbf{53.63}&\textbf{66.34}&49.56&77.72&\textbf{43.45}&30.78&38.29&76.45&67.21&68.80&62.32&62.22&49.37&50.19&89.11&59.02\\
CSENet~\cite{CSENet2024}
&70.24&74.99&49.77&40.15&72.16&42.26&58.75&55.39&77.93&33.70&35.13&33.58&\underline{77.36}&71.62&73.80&70.89&70.32&58.43&59.21&89.08&60.74\\
\midrule
Ours                                &\underline{72.87}&\underline{75.26}&\underline{53.28}&\underline{43.56}&73.91&47.60&63.31&\underline{56.20}&\underline{79.09}&37.21&\textbf{44.31}&\underline{39.38}&76.86&\underline{72.59}&\underline{73.88}&\underline{71.15}&\underline{71.50}&\underline{58.96}&\underline{59.48}&\underline{89.28}&\underline{62.98} \\
Ours$^{\dagger}$                   
&\textbf{73.53}&\textbf{75.99}&\textbf{55.46}&\textbf{45.21}&\underline{75.06}&\underline{49.59}&\underline{64.60}&\textbf{58.42}&\textbf{80.21}&\underline{38.02}&\underline{44.26}&\textbf{40.39}&\textbf{77.51}&\textbf{74.00}&\textbf{74.93}&\textbf{73.06}&\textbf{72.82}&\textbf{60.60}&\textbf{61.54}&\textbf{89.71}&\textbf{64.25} \\
\bottomrule
\end{tabularx}}
\vspace{-0.2cm}
\caption{
\textbf{Per-class IoU comparison in various settings on the validation set of LIP}. Here, $^{\dagger}$ means test time augmentation. The best values are marked in bold, and the second-best values are marked with an underline.
}
\label{tab:perclassIoU}
\vspace{-0.4cm}
\end{table*}

\subsection{Experimental Settings}
\noindent\textbf{Baseline Network:}
We employ the basic structure and network settings provided by ViT-Adapter~\cite{chen2022vitadapter} as the baseline to validate the effectiveness of our method. ViT-Adapter includes three modules: the spatial prior module, the spatial feature injector module, and multi-scale feature extractor module. The spatial prior module is used to capture initial local semantic information from the input image. The spatial injector module injects spatial features into the backbone, and the multi-scale feature extractor reconstructs multi-scale features from backbone and previous stage. The backbone of ViT-Adapter is a plain Vision Transformer (ViT). To take advantage of the merit of ViT-Adapter, we substitute our network for the plain ViT backbone of ViT-Adapter. Although ViT-Adapter can integrate the information of the backbone and local details, which is beneficial for dense prediction, its backbone needs to be trained from scratch, and the final performance highly determined by the learning ability of the backbone. Through utilizing our designed modules (SRM and FTM) simultaneously, the constructed backbone (SCHNet) that is based on pre-trained SAM and CLIP effectively combines the high semantic understanding ability of CLIP and fine-grained segmentation ability of SAM and significantly improves the human parsing performance in much shorter training time.    

\noindent\textbf{Implementation details of SCHNet:}
In the experiments, we employ the pre-trained Vision Transformer Base that is the ViT-B-16 model released by OpenAI as the CLIP backbone. The model consists of $12$ Transformer layers for the image encoder. We extracted the image patch tokens after each stage of the image encoder (i.e., layers $3$, $5$, $7$, and $11$) as the CLIP multi-level vision features. The normalized class-embedding from the feature of the last stage of CLIP is leveraged to calculate the similarity with textual feature from the CLIP text encoder. The pre-trained model (sam-vit-l) that has $24$ transformer layers acts as SAM image encoder. We extract the image patch tokens from layers $6$, $12$, $18$, and $24$ as multi-level image features that are ultimately input to the decoder. We utilize $UperNet$ ~\cite{uperNet-2018} the popular semantic segmentation decoder for our human parsing tasks. 

\noindent\textbf{Data Augmentation:}
In the training phase, standard augmentations are applied, such as mean subtraction, random scaling in the range of $[0.5, 2.0]$, photometric distortions and random left-right flipping. We randomly crop the large image or pad the small images into a fixed size for training (e.g., 480$\times$480 for LIP ,PPP, and CIHP ).

\subsection{Training}
We adopt ViT-Adapter~\cite{chen2022vitadapter} as the basic network structure, pre-trained CLIP~\cite{clip2021} as multi-modal network to extract strong semantic information and pre-trained SAM~\cite{sam2023} as the backbone to excavate semantic and image features to input the decoder of human parsing. After combining our proposed SRM and FTM modules, the whole network is trained for $60K$, $30K$, and $80K$ iterations on the LIP, PPP, and CIHP datasets, respectively. The weights of CLIP and SAM are frozen. We set the inserted SRM and FTM modules, patch-embedding block, the spatial prior module, the injector module, the multi-scale feature extractor, and UpperNet decoder tunable when training the whole network. The optimization process is conducted on $4$ NVIDIA $A5000$ GPUs using the AdamW optimizer with the initial learning rate of $6\times 10^{-4}$. We utilize a brief linear warming up learning of $1,500$ iterations with warming up ratio of $1\times 10^{-5}$. The batch size is set to $16$.

\subsection{Inference}
In the inference phase, the pixel accuracy (pixAcc), mean accuracy, and mean pixel Intersection-over-union (mIoU) are leveraged as the evaluation metrics for the LIP dataset, and the mIoU for the PPP and CIHP datasets. 
We average the predictions of the input and the flipped input to further improve the performance, and averaged predictions of multiple scaled inputs~\cite{XiaoPCNet2020,Ke2019CIHP} (e.g. $0.75$, $1.0$, $1.25$, and $1.5$ ) are also employed in our method to achieve a best performance on three well-known datasets. 

\subsection{The Experimental Results}
\label{sec:result}
To demonstrate the efficacy of our proposed methods, we conduct extensive experiments on LIP, PPP, and CHIP datasets. We leverage the cross-entropy and mIoU losses between the prediction result and the Ground Truth of input to obtain improved performance. The LIP validation dataset is utilized to evaluate the efficacy of each of our modules on human parsing. 

\noindent\textbf{Performance on LIP database:}
We showcase the performance comparison of the proposed method to other methods on the LIP validation set. As \cref{tab:perclassIoU} shows, our method achieves the highest performance on the majority of human parts in terms of mIoU, especially on hat, glove, sock, pants, scarf, and left-right human parts, which validates the efficacy of combining the strong semantic understanding ability of CLIP and the fine-grained segmentation capacity of SAM. The proposed SCHNet allows us to report new state-of-the-art performance (64.25\%) on LIP and the performance of our method is substantially higher ($3.51\%)$ than the previous best achievement, as summarized in \cref{tab:perclassIoU}. Our method achieves $3.03\%$, $5.06\%$, $5.69\%$, and $9.13\%$ higher performance in sock, glass, glove, and scarf parsing, which demonstrates the merit of using the SAM advantage in fine-grained segmentation of human parts. On the other hand, we boost the parsing result by $2.38\%$, $1.13\%$, $2.17\%$, $2.50\%$, $2.17\%$ and $2.33\%$ on the L/R arm, L/R leg and L/R shoe, respectively, which verifies the advantage of integrating strong semantic understanding ability of CLIP into our method. We observe significant improvements in average parsing performance compared to other methods, as reported in \cref{tab:example1}. For instance, we achieve $3.03\%$, and $5.23\%$ higher performance than previous methods (CSENet~\cite{CSENet2024} and DTML~\cite{DTML2024}) that utilize multi-stage feature maps to boost the performance. 
Our model is $3.23\%$ higher than HssN+~\cite{HssN2024} that exploited structured label constraints and structured representation learning, which supports our assumption that we can utilize the global understanding ability of CLIP to learn the latent structured representation of the human body. HssN+~\cite{HssN2024} trained the model on LIP dataset for 160K iterations, while our model is trained only for 60K iterations. As \cref{tab:example1} shows, we achieve the best performance in terms of Pixel Accuracy, Mean Accuracy, as well.

From the viewpoint of the training computational complexity, as shown in~\cref{tab:Cost}, what we should mention here is that we only take $17$ hours to train our model with $4$ NVIDIA $A5000$ GPUs (24GB VRAM), whereas, UniHCP~\cite{ci2023unihcp} took $120$ hours in total using $88$ NVIDIA $V100$ GPUs (32GB VRAM), which shows training efficiency while simultaneously achieving better accuracy on both the LIP and CIHP datasets. 

\begin{table}\centering
\resizebox{0.9\linewidth}{!}{%
\small
\begin{tabularx}{\linewidth}{ r c c c}
\toprule
Method    & Pixel Acc. & Mean Acc. & mIoU  \\ 
\midrule
CorrPM~\cite{Zhang2020CorrPM-8}                     & 87.68     &67.21    &  55.33\\
BGNet~\cite{Zhang2020BGNet-23}&-&-&56.82\\
ISNet~\cite{ISNet2021ICCV} &-&-&56.96\\
MCIBISS~\cite{MCIBISS2021ICCV} &-&-&56.99\\
PCNet~\cite{XiaoPCNet2020} & - & - &57.03\\
DTML~\cite{DTML2024}&\underline{89.34}&71.49&59.02\\
HHP~\cite{Wang2020HTypePRR-20} &89.05& 70.58&59.25 \\
SCHP~\cite{li2020self} &-&-&59.36\\
M2FP~\cite{yang2024humanparsing} & 88.93&-&59.86\\
CDGNet~\cite{LiuHwang2022} & 88.86     &71.49    &  60.30\\
HssN+~\cite{HssN2024} &-&-& 61.02\\
CSENet~\cite{CSENet2024} & 89.21     &\underline{72.98}    &  61.22\\
HumanBench~\cite{tang2023humanbench}&- & - & 62.90\\
UniHCP~\cite{ci2023unihcp} &- & - & \underline{63.86}\\
\midrule
Ours     & \textbf{90.04}     &\textbf{78.42}    &  \textbf{64.25}\\
\bottomrule
\end{tabularx}%
}
\vspace{-0.2cm}
\caption{\textbf{Comparison of different methods on the validation set of the LIP dataset}. 
The bold is the best one and the underline is the second best.}
\label{tab:example1}
\vspace{-0.2cm}
\end{table}

\begin{table}
    \centering    
    \resizebox{0.9\linewidth}{!}{
    \small
    \begin{tabularx}{\linewidth}{rccccc}
    \toprule
    Method & GPU & GPU & Training & \multicolumn{2}{c}{mIoU} \\
    & & Num. & Time & LIP & CIHP\\
    \midrule
    UniHCP~\cite{ci2023unihcp} & $V100$  & 88 & 120hr & 63.86 & 69.80\\
    Ours & $A5000$ & 4 & 17hr & \textbf{64.25} & \textbf{72.27}\\
    \bottomrule
    \end{tabularx}
    }
    \vspace{-0.2cm}
    \caption{\textbf{Training cost comparison} of foundation model-based human parsing methods.}
    \vspace{-0.4cm}
    \label{tab:Cost}
\end{table}

\noindent\textbf{Performance on PPP database:}
\cref{tab:example2} lists the detailed results on the Pascal-Person-Part test dataset. As seen, our method achieves the best performance in terms of mIoU. Remarkably, our method outperforms the previous method by significant margins. The performance of our model surpasses the previous best value by $2.11\%$, which is remarkable since this dataset only possesses a small number of training images. Moreover, we merely train our model for $30K$ iterations. The iteration number is much smaller than that of HssN+~\cite{HssN2024} ($80K$). Furthermore, we do not have to construct the class hierarchy, which improves the generality of our method. HssN+ ~\cite{HssN2024} provided the segmentation scores using the average results of multiple scales \{0.5,0.75,1.0,1.25,1.5,1.75\}, whereas we outperform it by only employing the multiple scales of \{0.75,1.0, 1.25, 1.5\}.

\begin{table}\centering  
\resizebox{0.9\linewidth}{!}{%
\begin{tabularx}{\linewidth}{r c c c}
\toprule
Method &Backbone& Pub.&mIoU \\ 
\midrule
DTML~\cite{DTML2024} &HyRNet&TPAMI$_{[2024]}$ & 71.93\\
M2FP~\cite{yang2024humanparsing} &ResNet101&IJCV$_{[2024]}$ & 72.54\\
SST~\cite{sst2023} &ResNet101&CVPR$_{[2023]}$&74.96\\
HssN+~\cite{HssN2024} &ResNet101& TPAMI$_{[2024]}$&\underline{76.56}\\
\midrule
Ours   &ViT-L&-&\textbf{78.67}\\
\bottomrule
\end{tabularx}%
}
\vspace{-0.2cm}
\caption{
\textbf{Comparison of mIoU on the Pascal-Person-Part test set}. The bold font is the best value and the underline is the second one. }
\label{tab:example2}
\vspace{-0.2cm}
\end{table}

\begin{table}\centering\normalsize
\resizebox{0.9\linewidth}{!}{%
\begin{tabularx}{\linewidth}{r c c c}
\toprule
Method&Backbone 
& mIoU \\ 
\midrule
PGN~\cite{Gong2018PGN} & DeepLabV2
&55.80\\
Graphonomy(PASCAL)~\cite{Ke2019CIHP}  & DeepLabV3+
&58.58\\
CorrPM~\cite{Zhang2020CorrPM-8}  &ResNet101 &60.18\\
SNT~\cite{Ji2020LSNT-22} &ResNet101& 
60.87\\
PCNet~\cite{XiaoPCNet2020} &ResNet101 
&61.05\\
CDGNet~\cite{LiuHwang2022}&ResNet101 &65.56\\
CSENet~\cite{CSENet2024} &ResNet101 &  67.41\\
HumanBench~\cite{tang2023humanbench}& ViT-L & 67.50\\
M2FP~\cite{yang2024humanparsing} &ResNet101&69.15\\
UniHCP~\cite{ci2023unihcp} &ViT-L& \underline{69.80}\\
\midrule
Ours & ViT-L & \textbf{72.27}\\
\bottomrule
\end{tabularx}%
}
\vspace{-0.2cm}
\caption{
\textbf{Performance comparison with state-of-the-art method on CIHP validation dataset}. The bold font is the best value and the underline is the second one. }
\vspace{-0.4cm}
\label{tab:CIHP}
\end{table}

\noindent\textbf{Performance on CIHP database:} ~\cref{tab:CIHP} provides the comparisons of our method against previous methods. We observe substantially higher performance ($2.47\%$) than the previous best result under much shorter training time and much fewer GPU resources. Note that the improvement on the multiple human parsing task is extremely challenging. Obviously, our method attains impressive results. We outperform CorrPM ~\cite{Zhang2020CorrPM-8} that employed keypoints of human body by $12.09\%$. Notably, we surpass the methods that utilized the semantic tree (SNT~\cite{Ji2020LSNT-22}), the distribution rule (CDGNet~\cite{LiuHwang2022}) of the human body, the pre-training on multiple human centric tasks (HumanBench~\cite{tang2023humanbench}) by $11.40\%$, $6.71\%$, $4.77\%$, respectively, which verifies the superiority of our SRM module in employing the latent knowledge of human body structure that lies in CLIP to improve the performance. Our method gives boosts of $4.86\%$ and $3.12\%$ compared to CSENet~\cite{CSENet2024} that utilized multi-stage feature maps for accurate parsing and M2FP that exploited a heavier transformer decoder and a larger image size for high performance, respectively, which confirms the efficacy of our FTM module that tunes SAM to leverage fine-grained segmentation advantages.

\begin{table}[]\centering\small
\resizebox{1.0\linewidth}{!}{%
\begin{tabular}{c c c c|c c c } 
\hline
\multicolumn{4}{c|}{Method} & \multicolumn{1}{c}{\multirow{2}{*}{Pixel Acc}} & \multirow{2}{*}{Mean Acc} & \multirow{2}{*}{mIoU} \\ 
CLIP   & SAM   & SRM   & FTM  & \multicolumn{1}{c}{}   &  &      \\ \hline
\checkmark&-&-&-&86.09&68.01&51.89\\ 
- &\checkmark&-&-&87.20&69.49&54.61  \\ 
\checkmark&\checkmark&\checkmark&-&89.27&76.86&61.85\\
\checkmark&\checkmark&\checkmark&\checkmark&89.54&78.12&62.98\\ \hline
\end{tabular}
}
\vspace{-0.2cm}
\caption{
\textbf{Each component of our method} is evaluated on the LIP validation set including CLIP, SAM, SRM and FTM modules.}
\label{tab:Ablation}
\vspace{-0.4cm}
\end{table}

\subsection{Ablation studies.} 
As shown in~\cref{tab:Ablation}, directly employing CLIP and SAM as the backbones results in relatively low performance. SAM outperforms CLIP by $2.72\%$ mIoU, indicating that detailed spatial information plays a more critical role in human parsing. We assume that this is because certain human parts are small and are located in the extremities of the body, requiring precise localization. However, semantic information is also essential for classifying human parts, recognizing larger-scale patterns, and distinguishing between left and right body parts. By integrating CLIP into SAM through the SRM module, we observe a significant performance improvement of $7.24\%$, $7.37\%$, and $2.07\%$ in terms of mIoU, Mean Accuracy, and Pixel Accuracy, respectively. Furthermore, an additional performance gain of $1.13\%$ in mIoU is achieved by applying the FTM module, which introduces learnable tokens and fine-tunes the feature maps to better adapt the VFM models to the human parsing task. These results clearly demonstrate the effectiveness of our proposed modules in fully leveraging the knowledge embedded in pretrained vision foundation models for accurate human parsing.

\section{Conclusion}
In this paper, we propose two efficient modules to fully leverage the strengths of pre-trained CLIP and SAM for human parsing. While CLIP provides strong semantic understanding but lacks detailed localization, SAM excels at fine-grained segmentation but struggles with semantic-aware parsing, often leading to over-segmentation in regions of the same semantic category. To address these limitations, we propose SRM, which injects multi-level semantic features from CLIP into SAM to enhance the semantic understanding of feature maps. Additionally, we design FTM, which incorporates learnable tokens to adapt general features to the human parsing domain, enabling efficient fine-tuning with reduced training time while preserving spatial details and semantic consistency. By combining these modules, our method achieves fast convergence and state-of-the-art performance on standard human parsing benchmarks.

{
    \small
    \bibliographystyle{ieeenat_fullname}
    \bibliography{main}
}

\end{document}